\documentclass{article}
\usepackage[preprint]{neurips_2023}

\usepackage{hyperref}
\hypersetup{
    colorlinks = true,
    citecolor = {blue},
    urlcolor = {blue},
}

\usepackage{graphicx}
\usepackage{amsmath}
\usepackage{natbib}
\setcitestyle{numbers,square}
\usepackage[utf8]{inputenc} 
\usepackage[T1]{fontenc}    
\usepackage{hyperref}       
\usepackage{url}            
\usepackage{booktabs}       
\usepackage{amsfonts}       
\usepackage{nicefrac}       
\usepackage{microtype}      
\usepackage{xcolor}         

\title{Are Human Conversations Special?\\A Large Language Model Perspective}
\author{
    Toshish Jawale \\
    Symbl.ai\\
    \texttt{toshish@symbl.ai}\\
\And
    Chaitanya Animesh \\
    Symbl.ai\\
    \texttt{chaitanya.animesh@symbl.ai}\\
\And
    Sekhar Vallath \\
    Symbl.ai \\
    \texttt{sekhar.vallath@symbl.ai}\\
\AND
    Kartik Talamadupula \\
    Symbl.ai\\
    \texttt{kartik.t@symbl.ai}\\
\And
    Larry Heck \\
    Georgia Institute of Technology\\
    \texttt{larryheck@gatech.edu}
}

\date{March 2024}

\begin{document}

\maketitle

\begin{abstract}
This study analyzes changes in the attention mechanisms of large language models (LLMs) when used to understand natural conversations between humans (human-human). We analyze three use cases of LLMs: interactions over web content, code, and mathematical texts. By analyzing attention distance, dispersion, and interdependency across these domains, we highlight the unique challenges posed by conversational data. Notably, conversations require nuanced handling of long-term contextual relationships and exhibit higher complexity through their attention patterns. Our findings reveal that while language models exhibit domain-specific attention behaviors, there is a significant gap in their ability to specialize in human conversations. Through detailed attention entropy analysis and t-SNE visualizations, we demonstrate the need for models trained with a diverse array of high-quality conversational data to enhance understanding and generation of human-like dialogue. This research highlights the importance of domain specialization in language models and suggests pathways for future advancement in modeling human conversational nuances.
\end{abstract}

\section{Introduction}
Understanding natural language is a cornerstone of artificial intelligence, with transformer-based large language models (LLMs) representing a significant leap forward in this endeavor~\cite{vaswani2017attention,minaee2024large}. These models have shown remarkable proficiency across a range of linguistic tasks, yet their performance varies widely across different types of data. Domain-specialized LLMs have shown greater effectiveness than general LLMs in various specialized settings such as code \cite{rozière2024code, li2023starcoder}, math \cite{azerbayev2024llemma}, finance \cite{wu2023bloomberggpt}, and medical \cite{labrak2024biomistral, nazi2023large} domains. However, there has been less focus on natural human-human conversations, which embody a rich tapestry of nuances, contexts, and unspoken cues \cite{tur2011human}. We perform a comprehensive analysis of how transformer-based LLMs -- specifically focused on the LLaMa-2 13b \cite{touvron2023llama2} model -- process and interpret human conversations as opposed to other data such as web content, code, and mathematical texts. 

Formal "textbook" conversations such as those taught in classroom settings to analyze conversational structures do not typically exhibit the same characteristics as speakers engaged in speaking and communicating naturally~\cite{rings1986authentic}. Spoken conversations are spontaneous, and to operate effectively in conversations, the knowledge of the participating entity has to stretch far beyond the awareness of sounds and words in terms of extrinsic and intrinsic knowledge. As a result of years of evolution and a social environment where the use of language in conversation is practiced throughout each day, humans can structure and build conversations appropriate to the situation without any formal training, and adapt to changing norms with time~\cite{pridham2013language}. These emergent traits are not prevalent in documents, articles, or social forums which constitute a large portion of web data; or in other domain-specific corpora like code.

Recent years have seen significant progress in modeling conversational interactions using dialogue state tracking methods~\cite{zhong-etal-2018-global, wu2019transferable, lee2021dialogue, hu2022context, 10147329}. However, these efforts focus on two-party interactions where one party is human and the other is a machine (human-machine). These interactions are typically short with only a few conversational turns. In contrast, multiparty human-human interactions typically contain a significantly higher number of turns; much higher complexity in contextual dependencies; more sophisticated emotional depth; can have more than two parties involved; and all parties are human. This lack of attention is also one of the limiting factors in making human-machine interactions more natural, and modeling a deeper understanding of humans' social and emotional behaviors can improve interaction with humans~\cite{doi:10.1177/0278364921990671} in 1-on-1 as well as multi-party settings~\cite{brennan1991conversation}.

In this work, we begin by analyzing the proportion of authentic human-human conversations in the web data used to pretrain state-of-the-art LLMs. Our analysis finds that authentic human conversations are rare in occurrence on the web, and the vast majority of ``conversation data'' merely refers to textbook conversations, creating a huge data scarcity for authentic human conversations. 

Our investigation centers on three key aspects: attention distance, dispersion, and interdependency within different data domains. Through quantitative analysis of attention entropy and qualitative inspections of attention patterns, we seek to understand the intricacies of model behavior across domains. Moreover, we employ t-SNE visualizations to compare the hidden state representations of language models when exposed to different types of data, allowing us to visually assess how domain-specific characteristics are encoded within the models, offering insights into their ability to distinguish and adapt to varied linguistic challenges.

In doing so, we aim to shed light on the current limitations of generalist models in handling conversational data and argue for the development of more domain-specialized models. By highlighting the unique demands of human-human conversation on LLMs, our research underscores the need for incorporating diverse and authentic conversational data into model training processes. This, we believe, is crucial for advancing the capabilities of LLMs in understanding and generating human-like dialogue, thereby paving the way for more natural and effective human-AI interactions.

\section{Human-Human Conversations}

The majority of human-human conversations are conducted in spoken language rather than via written texts. Natural human conversation is an interactive exchange between two or more people, with a format that can be one-on-one or between multiple people. Examples of such interactions include chats between family or friends, at work, or in the public domain; and can be conducted either face-to-face or virtual, with or without visual elements in it. Conversations, however, are far more than just the words that they are made up of~\cite{pridham2013language}. The textual representation of a spoken conversation loses significant information from the speech and visual channels/modalities. Speech contains information about the speaker in terms of their emotions, intelligence, age, psychological traits, and various other properties~\cite{spirina2016human}. The combination of information in visual and speech channels is manifested through body language and gestures and their intensities; and prosodic features such as speed, intonation, speed, amplitude, silence, and laughter. However, the textual representation of spoken language contributes primarily to the meaning and knowledge of the thought in the exchange, while indirectly modeling subtle cues from the speech and visual modalities. Understanding conversation in its complete sense requires understanding the purpose behind the words and the situational, emotional, social, and contextual understanding established in the conversation and their evolution until a specific point in the conversation~\cite{pridham2013language}.

\subsection{Characteristics of Human-Human Conversations}

Human-human conversations are distinguished by several key characteristics:

\textbf{Interactivity:} Unlike static web data, human conversations are highly interactive, with participants actively responding to and building upon each other's contributions. This interactivity involves turn-taking, feedback signals (e.g., nodding, "uh-huh"), and adjustments in discourse based on the other participants' responses. Such dynamics are absent in non-conversational data, where the information flow is usually unidirectional.

\textbf{Contextuality:} Conversations are deeply embedded in specific contexts, which include physical surroundings, social relationships, cultural backgrounds, and the participants' shared history. This context influences not only the content but also the form of the conversation, including language choice, tone, and register. In contrast, domains like code or mathematics are characterized by a high level of abstraction, process, and standardization, where context plays a minimal role in the interpretation of the data.

\textbf{Adaptability:} Participants in a conversation continually adjust their speech based on immediate feedback from their interlocutors. This adaptability covers a wide range of aspects, from changing topics smoothly to modifying speech patterns for clarity or emphasis. Such dynamic adjustments are specific to human interactions and are not found in structured data domains like code, where the syntax and semantics follow rigid, predefined rules.

\textbf{Emotional and Psychological Dimensions:} Conversations convey not just factual information but also emotional and psychological states. Through tone, pace, volume, and choice of words, speakers can express a wide range of emotions and attitudes. These nuanced emotional layers add depth to human conversations that are typically absent in other data types, where emotional expressiveness is either irrelevant or vastly simplified.

\subsection{Differences from Other Web Data and Domains}

Compared to other forms of web data (e.g., blogs, forums, social media posts), human-human conversations present unique challenges and opportunities for analysis. While web data can be rich in content, it often lacks the immediacy and interactivity of live conversations. Even interactive web platforms like forums and comment sections do not replicate the dynamic, back-and-forth nature of spoken exchanges. Human-human conversations on the other hand are not only multi-turn and multi-party, but also have a natural style of communication. They are a medium for expression and not merely fact exchange, where participants can often interrupt others, subtly change the context of conversation, and use non-verbal cues to shape the flow of interaction. The lack of real-time feedback mechanisms in web forums and social media platforms results in a different kind of interaction, one that is less about dialogue and more about individual expression and asynchronous responses.

Furthermore, when comparing conversations to domains like code or mathematics, the contrast becomes even more pronounced. Code and mathematical expressions are governed by strict logical structures and rules, offering little room for ambiguity or personal expression. The clarity and precision required in these fields stand in stark contrast to the fluidity and ambiguity often present in human conversations. Human conversations thrive on implicit meanings, cultural references, and shared knowledge, aspects that are largely irrelevant or minimized in coding and mathematical contexts.

\subsection{Human-Human Conversation Data on the Web}

The majority of the content on the internet is in the form of articles, documents, blogs, and forums where information is structured. Authentic human conversations are drastically less in proportion to written content in the web data. It has been challenging to find authentic human-human conversation data publicly that can be used for training models due to copyright, privacy, and intellectual property concerns. We analyze the web data from CommonCrawl\cite{CommonCrawl2023} dumps for human conversation data and the types of conversations in Table~\ref{table:commoncrawl-conversation-dist}. We randomly sample a subset of the dump and deduplicate it so that it can be used to approximate the data distribution between human conversations versus the rest of the data. We fine-tune a BERT~\cite{DBLP:journals/corr/abs-1810-04805} model for document classification using $\sim$194K samples containing human conversations and non-conversational documents in equal amounts. We find that natural human conversations are rare in the web domain: accounting for the $+0.0043\%$ error rate of the model, such conversations only account for a maximum of $\approx0.0128\%$ of the total data.\label{conversation-data-in-web}

\begin{table}[!ht]
  \centering
  \begin{tabular}{lll}
    \toprule
    \cmidrule(r){1-2}
    Type     & Percentage   & Err. \\
    \midrule
    Written/Documents & $99.9915\%$   & $\pm0.0043\%$     \\
    Human Conversations     & $0.00849\%$    & $\pm0.0043\%$     \\
    \bottomrule
    \smallskip
  \end{tabular}
  \caption{Distribution between Human Conversations and Written/Document data in CommonCrawl.}
  \label{table:commoncrawl-conversation-dist}
\end{table}

\section{Related Work}
Recent studies have shown the intricate ways in which various models, including Transformers and recurrent neural networks, encode dependency relations within texts~\cite{hewitt2019structural, raganato-tiedemann-2018-analysis}. Transformer models have been found to most effectively capture dependency relations within their middle layers~\cite{liu2019linguistic}.

Analysis of the attention distance within decoder-only transformer models~\cite{vig2019analyzing} has provided evidence supporting the hypothesis that deeper layers capture longer-distance relationships. This is a measurement of the mean distance spanned by attention for each head; and is calculated as the average distance between token pairs in all samples in the dataset, weighted by attention between the tokens:

\begin{equation} \label{eq:attention-distance}
\overline{D}_{\alpha} = \frac{\sum_{x \in X} \sum_{i=1}^{|x|} \sum_{j=1}^{i} \alpha_{i,j}(x) \cdot (i - j)}{\sum_{x \in X} \sum_{i=1}^{|x|} \sum_{j=1}^{i} \alpha_{i,j}(x)}
\end{equation}

The exploration of attention dispersion and entropy as measures of how attention is distributed across tokens offers additional insights into the mechanisms through which models understand and process patterns in language:

\begin{equation} \label{eq:entropy}
\text{Entropy}_{\alpha}(x_i) = -\sum_{j=1}^{i} \alpha_{i,j}(x) \log(\alpha_{i,j}(x))
\end{equation}

This body of work sets a context for our investigation into the unique characteristics of human-human conversations, comparing these dynamics against the backdrop of general web corpora, including articles, blogs, forums, and specialized domains such as mathematics and programming. Understanding the nuances of how models encode dependency relations and manage attention across different types of text is crucial in distinguishing the specifics of human conversational patterns.

Finally, the analysis conducted in this paper is similar in spirit to the work in the mutlilingual (large) language model (MLLM) community on the effect of using models trained on higher resource languages and datasets with data from lower resource settings. In one effort~\cite{joshi2020state}, the authors identify the lack of linguistic diversity when training models -- similar to the lack of diversity in data type when training LLMs, which is the focus of our present study; while in another~\cite{rust2021good}, a detailed empirical analysis is provided to show the differences between different languages. We take inspiration from these efforts for our attention-centric study of language models and the content used to train them.

\section{Analysis}
We explore the nuanced landscape of natural human conversations in comparison to other data domains. We use combined web data, code, and mathematics as data domains to assess the implications of these nuances for the understanding and development of large language models (LLMs).

\subsection{Attention Distance Difference}
Analyzing the difference between the attention distances as defined in Equation~\eqref{eq:attention-distance} between two domains can provide a better way to gain insights into how language models form relationships, especially longer-distance relationships in deeper layers by focusing on the difference between attention distances.

Given two domains, \(D_1\) and \(D_2\), with their respective sets of texts \(X^{D_1}\) and \(X^{D_2}\), the attention distance for each domain is calculated as:
\begin{equation} \label{eq:attention-distance-difference}
\overline{D}_{\alpha}^{D_k} = \frac{\sum_{x \in X^{D_k}} \sum_{i=1}^{|x|} \sum_{j=1}^{i} \alpha_{i,j}(x) \cdot (i - j)}{\sum_{x \in X^{D_k}} \sum_{i=1}^{|x|} \sum_{j=1}^{i} \alpha_{i,j}(x)}
\end{equation}

for \(k = 1, 2\), where \(\alpha_{i,j}(x)\) is the attention weight from token \(i\) to token \(j\) in text \(x\), and \(|x|\) is the length of text \(x\).

The difference in attention distance between the two domains can then be defined as:

\begin{equation} \label{eq:delta-d}
\Delta \overline{D}_{\alpha} = \overline{D}_{\alpha}^{D_1} - \overline{D}_{\alpha}^{D_2}
\end{equation}

This measure, \(\Delta \overline{D}_{\alpha}\), quantifies the difference in how attention spans across tokens vary between the two domains, providing insights into the structural differences in how information is processed and dependencies are captured in texts from \(D_1\) compared to \(D_2\).

By analyzing \(\Delta \overline{D}_{\alpha}\) we can find insights into domain specificity in transformer models by understanding how transformer models adapt their attention mechanism to structural and contextual differences between various domains. We can also identify if models tend to focus on closer or more distant token relationships when dealing with human-human conversations as opposed to more structured and document-oriented content. Positive values of difference in attention distance indicate that the attention distance in the second domain is longer than the first domain, whereas negative values indicate that it is shorter. Positive differences in the middle and end layers indicate more complex relationships requiring longer dependencies, and positive differences in the initial layers indicate longer syntactic and semantic relationships in the sequence tokens.

\subsection{Attention Dispersion}
We also calculate the entropy of the attention distribution based on Equation~\eqref{eq:entropy} to measure the attention dispersion. This provides insights into how domain-specific characteristics and the model's training influence its learning and processing strategies. High entropy is not always desirable, as it is indicative of a lack of focus or understanding. Similarly, very low entropy might suggest overfitting to specific tokens or phrases, potentially reducing the model's ability to generalize across varied inputs within the domain. We perform a comparison of attention dispersion between domains to understand the robustness of the model's understanding of a domain.

\textbf{High Entropy Domain:} A higher entropy in the attention distribution of a domain means that the model finds the information in that domain more uniformly informative or relevant, without specific tokens or phrases standing out as significantly more important than others. This suggests that the domain is more complex or less familiar to the model, leading it to distribute its attention more evenly rather than clearly identifying key information. This would also indicate more variety and ambiguity in how information is presented, requiring broader focus to capture the necessary context for understanding.

\textbf{Low Entropy Domain:} Domains with lower entropy in the attention distribution indicate that the model is focusing its attention more narrowly on specific tokens. This suggests that the model has learned to identify key tokens or phrases that are particularly informative or relevant for understanding or performing tasks in such a domain. The domain is more structured or contains clearer cues that the model can exploit to make predictions or understand content. It also reflects a higher level of familiarity or specialization of the model in this domain, allowing it to more effectively pinpoint the most relevant information.

\subsection{Interdependency Analysis}

Analyzing interdependencies between various aspects of text from different domains provides insights into underlying structures, patterns, and dynamics of information in domains such as communication, written documents, and code. We devise a novel ttmetric -- the {\em Interdependency Factor} (IF) -- to quantify the degree of interdependency between various \textit{aspects} of the data, as long as they can be modeled in a graph as nodes along with directed edges between them. This is intended to indicate the overall complexity of the domain along specific aspects. \textit{Aspect} in this context refers to any derived or absolute representation in a sequence. In this analysis, we use a tokenized representation of text in the domain. However, it can be useful to use higher-level segmentation such as themes that can be common across different domains. When this is not possible or the dependency is modeled at more granular levels by systems such as language models, a lower-level representation (such as tokens) can be used, where the weights on the edges are attention values~\cite{vig2019analyzing} between the tokens modeled by the transformer~\cite{vaswani2017attention} language model layers.

The Interdependency Factor (IF) is defined as follows:

Given a dataset of text samples, a set $N$ represents all identified node candidate labels in the graph, where each node $n_i \in N$ represents a distinct \textit{aspect} value. To analyze the interdependencies among these nodes, we construct a directed graph $G = (V, E)$, where $V$ corresponds to the set of vertices, with each vertex representing a node in $N$, and $E$ represents the set of directed edges between these vertices. Each edge $(n_i, n_j) \in E$ is associated with a weight $w_{ij}$, quantifying the strength or frequency of the transition or relationship from node $n_i$ to node $n_j$.

The adjacency matrix $A$ of graph $G$ is defined such that each element $a_{ij}$ within $A$ corresponds to the weight $w_{ij}$ of the edge from $n_i$ to $n_j$. The Interdependency Factor IF is then defined as follows:

\begin{equation}
IF = \frac{1}{|N|^2 - |N|} \sum_{i=1}^{|N|} \sum_{j=1, j \neq i}^{|N|} a_{ij}
\end{equation}

This calculates the $IF$ by averaging the weights of all directed edges in the graph, excluding self-transitions (where $i = j$). The normalization factor, $|N|^2 - |N|$, accounts for the total number of possible directed transitions between different nodes, ensuring that the $IF$ remains a relevant measure of interdependency across datasets of varying sizes and complexities. In cases where the weights are not available or not computable, $0$ and $1$ should be used to indicate the absence and existence of a dependency between two aspects.

\section{Experimental Setup}

\subsection{Domain Datasets}
In our analysis, we focus primarily on English data across the domains of human-human conversation, web, and math. Code data is randomly sampled across a variety of programming languages. We use $1000$ samples from each domain in our analysis.

\textbf{Human-Human Conversations:} In our study, a wide range of real-life natural conversations between humans across various business and casual settings is used. Scripted conversations such as movie scripts, and single-person presentations or talks are not included. Key aspects such as context dependencies, emotional expressiveness, idiomatic usage, and integration of general and localized or private knowledge are the focus. The data used for human conversations is a set of real conversations between people, processed using the Symbl.ai\footnote{\url{https://symbl.ai}} platform and anonymized by replacing PII and PCI information with synthetic data.

\textbf{Web Data:} Data from the internet containing various types of content such as blog posts, news articles, forums, social media content, etc. is generated using a randomly sampled subset from the CommonCrawl~\cite{CommonCrawl2023}. We perform a preliminary data cleanu[] to remove unnecessary HTML tags, and deduplicate to ensure the entries are unique.

\textbf{Code:} Source code from various programming languages, each with unique syntax and semantic frameworks, is used. The focus is on the structure and logic expressed in code, which contrasts with the unstructured and mostly informal nature of human-human conversations. The code data is curated from the GitHub dataset~\cite{githubcode2022}.

\textbf{Mathematics:} Mathematical expressions, problems, and proofs across different fields of mathematics make up this domain, highlighting the abstract, precise, and symbolic characteristics of mathematical communication from the Proof Pile 2 corpus~\cite{azerbayev2023llemma}.

\subsection{Language Model} \label{language-model-llama2}

We use a pretrained decoder-only transformer language model -- LLaMa-2 13b \cite{touvron2023llama2} -- for analyzing attention patterns and embeddings at various layers and heads. This model's architecture contains $40$ layers and $40$ attention heads. Although exact details of the LLaMa-2 model are not indicated in the accompanying technical report~\cite{touvron2023llama2}, the model was trained on data that is similar to the LLaMa-1 models~\cite{touvron2023llama}. This enables our assumption that the model was trained on approximately 82\% of data from web dumps from CommonCrawl and C4, 4.5\% of data from the code domain from GitHub, and 2.5\% of data from ArXiv, which consists of scientific data with some overlap with math. Apart from the data in the web corpus, the rest of the data is distributed between Wikipedia, books, and StackExchange corpora. After adjusting for the web corpus distribution based on our earlier analysis (c.f. Section~\ref{conversation-data-in-web}), the pretraining data of the model is expected to have between $\approx0.0069618\%$ and $\approx0.010496\%$ of human-human conversations. We choose LLaMa-2 13b because of its relatively smaller size, and since it has shown considerably good performance. This allows for the assumption that this model has a robust understanding of data from the web, code, and math domains; and it has also seen human-human conversation data from the web corpus, albeit in a very small amount.
\section{Results}
\subsection{Attention Distance Difference Analysis}
\label{subsec:attention_distance_diff_analysis}

We calculate the mean difference in attention distances \(\Delta \overline{D}_{\alpha}\) (c.f. Equation~\eqref{eq:delta-d}) for each of the human-human conversations, code, and math domains; and compare each of these in turn against general web data.

\begin{figure}[!ht]
    \centering
    \includegraphics[width=1\linewidth]{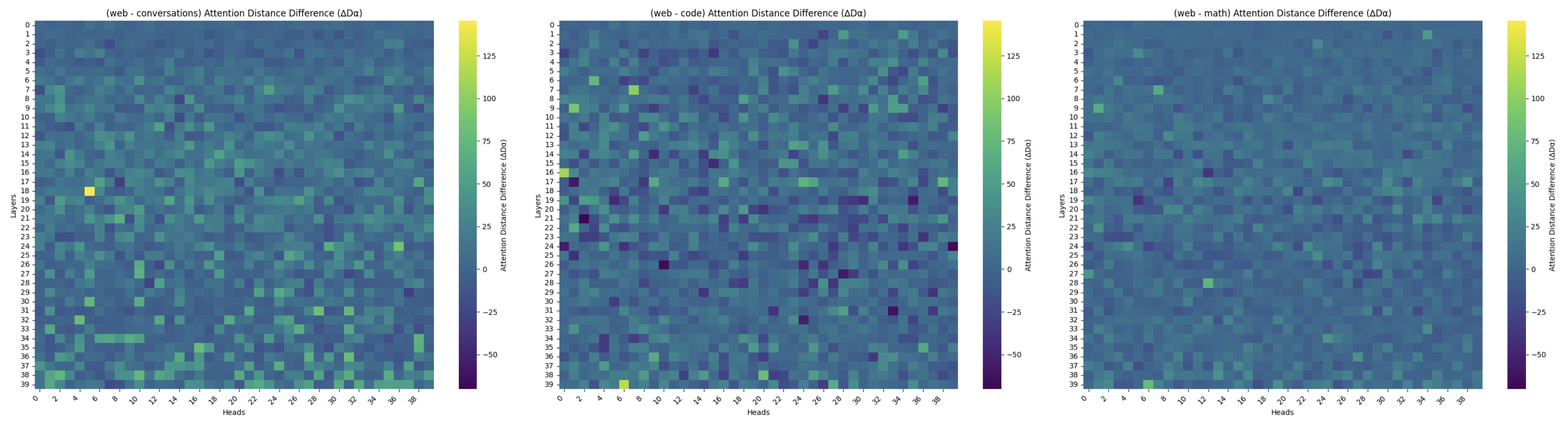}
    \caption{Heatmap of the Attention Distance Difference matrix (\(\Delta \overline{D}_{\alpha}\)) calculated with the first domain fixed as web data, and the second domain as human-human conversations, code, and math respectively. Higher values of difference in attention distances in deeper layers (left) indicate that human-human conversations demand deeper modeling of long-term contextual relationships than general web data. The comparison with code (center) indicates higher distances in the first half of the layers, but lower values in deeper layers, indicating more long-term relationships in structural aspects, but localized contextual relationships. However, the attention distances are quite spread across math (right) in comparison to the web.}
    \label{fig:attention-distance-difference-comparison}
\end{figure}

Figure~\ref{fig:attention-distance-difference-comparison} shows heatmaps of the Attention Distance Difference by layer (Y-axis) and head (X-axis), with one domain fixed as human-human conversations, code, and math respectively; and the other domain as general data from the web. We find significant differences in attention distances in deeper layers when comparing human-human conversations to web data. Higher values in these layers indicate that human conversations necessitate more robust modeling of long-term contextual relationships than general web corpora. This is consistent with the nature of human dialogue, where the flow of information often spans across several exchanges, requiring the model to maintain context over extended sequences. The comparison with code displays a distinctive pattern where higher attention distances are observed in the initial half of the layers, suggesting that models capture structural dependencies effectively in these stages. However, as we progress into deeper layers, there is a reduction in these values, which suggests that attention becomes more localized, focusing on closer contextual relationships. This reflects the structural and syntactic rigidity inherent in programming languages, where local context is often sufficient for understanding many dependencies. Mathematical texts exhibit a relatively even distribution of attention distances across layers when compared to web data. This implies that mathematical texts, with their symbolic and formulaic nature, require a balanced approach where both local and long-distance relationships are equally pertinent across all layers of the model. 

\begin{figure}[!ht]
    \centering
    \includegraphics[width=1\linewidth]{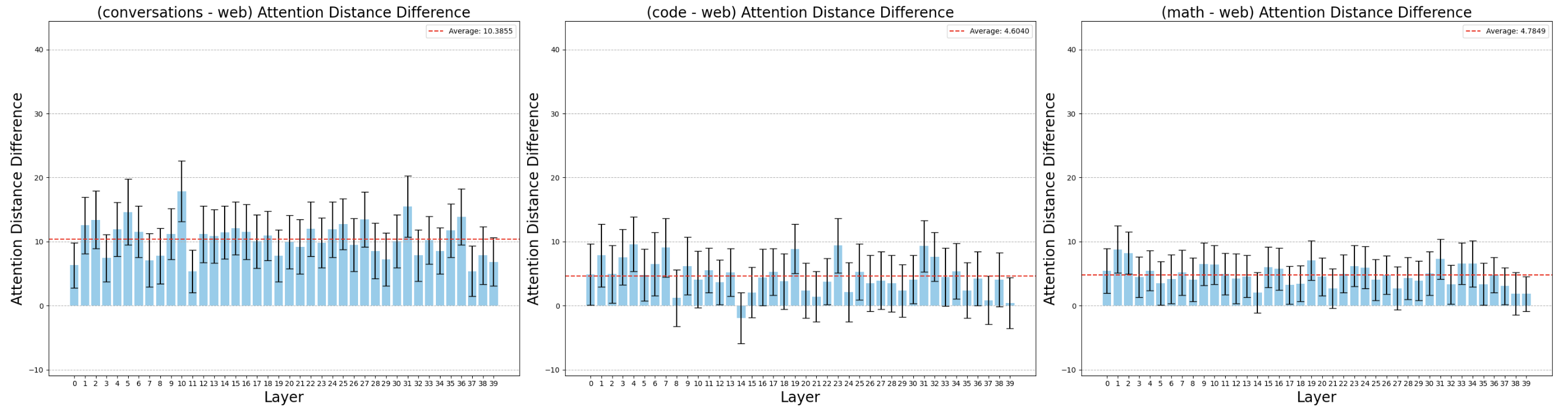}
    \caption{Attention Distance Difference by Layer across all heads calculated with the first domain as web data, and the second domain as human-human conversations (left), code (middle), and math (right) respectively.}
    \label{fig:attention-distance-difference-by-layer-comparison}
\end{figure}

\begin{figure}[!ht]
    \centering
    \includegraphics[width=1\linewidth]{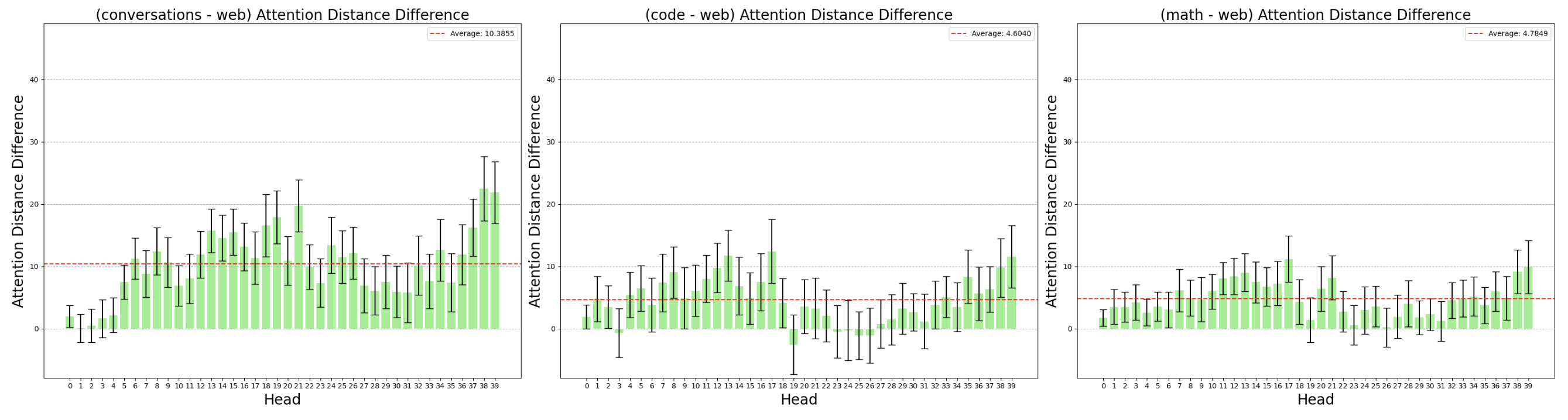}
    \caption{Attention Distance Difference by Head across all layers calculated with the first domain as web data, and the second domain as human-human conversations (left), code (middle), and math (right) respectively.}
    \label{fig:attention-distance-difference-by-head-comparison}
\end{figure}

\begin{table}[!ht]
  \centering
  \begin{tabular}{lll}
    \toprule
    \cmidrule(r){1-2}
    \(D_1\)     & \(D_2\)  & \(\Delta \overline{D}_{\alpha}\)   \\
    \midrule
    Human-Human Conversations & Web       & $10.3855$    \\
    Code & Web      & $4.6040$   \\
    Math & Web     & $4.7849$    \\
    \bottomrule
    \smallskip
  \end{tabular}
  \caption{Average attention distance difference between human-human conversations, code, and math domains with web data. A higher value indicates longer contextual dependencies.}
  \label{table:average-attention-distance-difference-by-domain}
\end{table}

Almost all the layers exhibit approximately equal differences in attention, with lower differences manifesting in the final layers (which are typically optimized for generation) as seen in Figure~\ref{fig:attention-distance-difference-by-layer-comparison}. Differences in the initial layers are typical across all domains, as syntactical and semantic modeling representations of the model are different across domains. However, more complex relationships are modeled in the middle layers, where we see significantly higher differences for conversation-web, as compared against code-web and math-web pairs. When compared by individual head in Figure~\ref{fig:attention-distance-difference-by-head-comparison}, the initial heads show very little difference from the web domain; but the middle heads and heads towards the end exhibit significant deviation from the web domain. These differences in the middle heads are less pronounced in the code and math domains, which indicates that human-human conversation domain modeling tends to have higher attention distances across most heads when compared to code and math domains.

This effect also calls to mind a similar effect in the domain of multilingual large language models (MLLMs): namely, the differences in how the same model (and tokenizer) handles different languages. The work of ~\citet{joshi2020state} points out the lack of linguistic diversity when training models that are intended for use in multilingual settings. This results in several problems, one of which concerns the issue of tokenization of text. Specifically, \citet{rust2021good} show that for two key metrics -- {\em sub-word fertility} and {\em proportion of continued words} -- lower represented languages tend to do worse than languages with higher representation (such as English). It is quite possible that a similar effect manifests when dealing with different kinds of data that large language models are trained on: data that tends to be represented poorly (e.g. human-human conversations) also tends to exhibit a higher attention distance difference. We are currently actively studying this phenomenon for data domains used for language model training.

\subsection{Attention Dispersion Analysis}

\begin{figure}[!ht]
    \centering
    \includegraphics[width=1\linewidth]{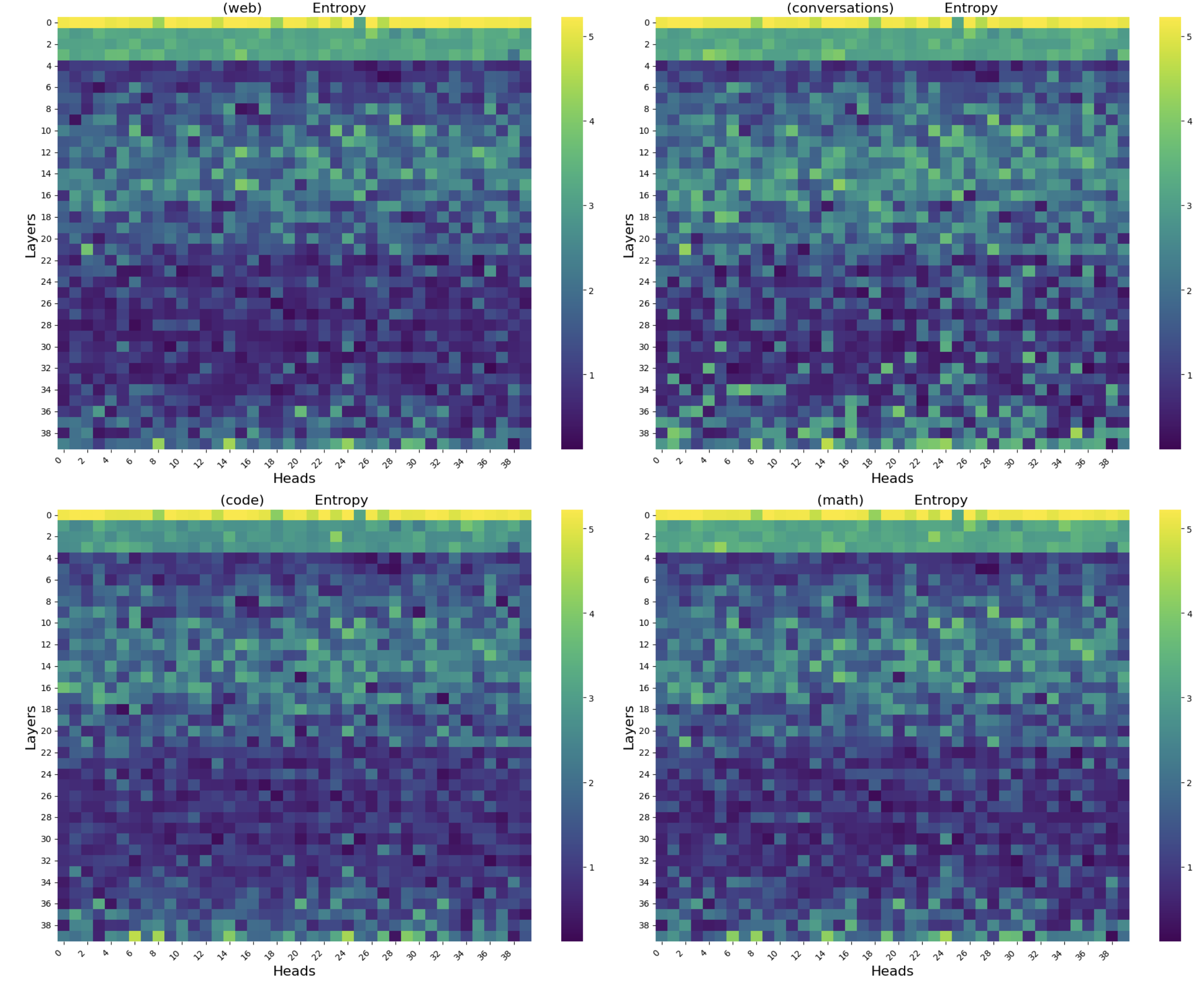}
    \caption{Heatmap of mean attention entropy for web, human-human conversations, code, and math domains respectively. Higher values indicate more attention diffusion. Human-human conversations show the highest diffusion in attention, especially in the middle and end layers. Attention diffusion in web, code, and math domains is similar, with small differences.}
    \label{fig:entropy-all-domains}
\end{figure}

\begin{figure}[!ht]
    \centering
    \includegraphics[width=1\linewidth]{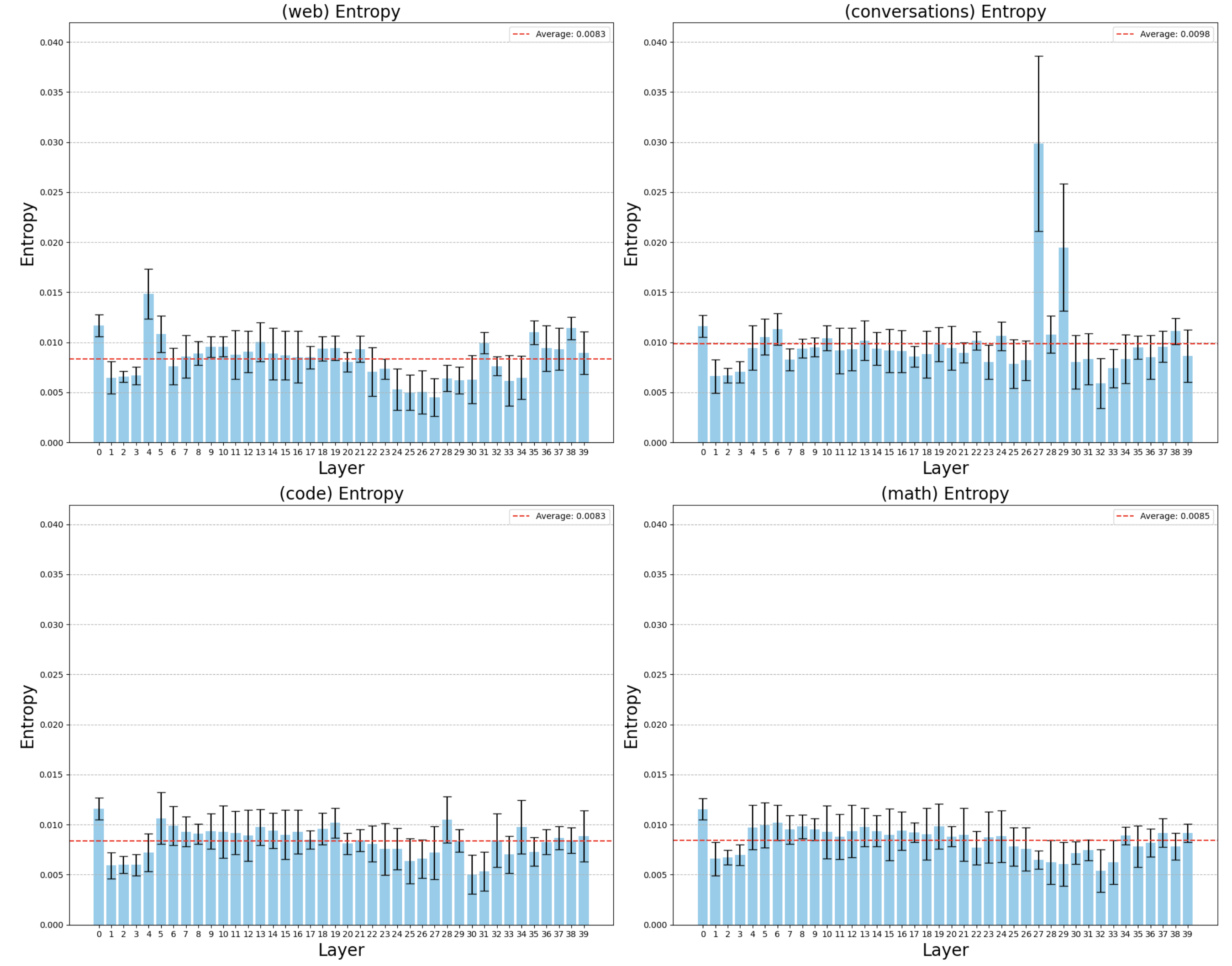}
    \caption{Mean attention entropy by layer across all heads with first token attention removed for web, human-human conversations, code, and math domains respectively. Higher values indicate more attention diffusion in the layer.}
    \label{fig:entropy-by-layer-excl-first-tok}
\end{figure}

\begin{figure}[!ht]
    \centering
    \includegraphics[width=1\linewidth]{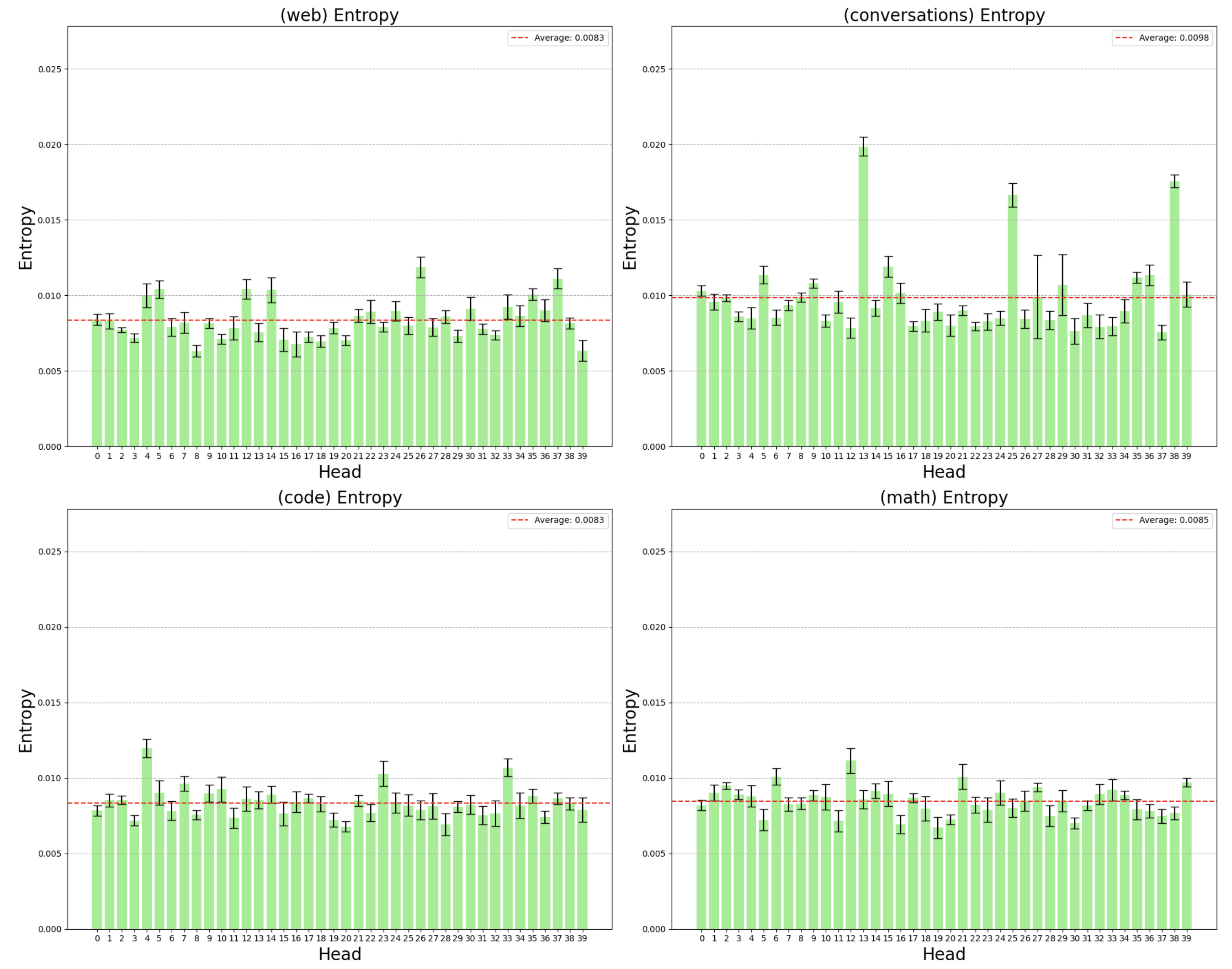}
    \caption{Mean attention entropy by the head across all layers with first token attention removed for web, human-human conversations, code, and math domains respectively. Higher values indicate more attention diffusion in the head.}
    \label{fig:entropy-by-head-excl-first-tok}
\end{figure}

To study the dispersion of attention across domain data, we calculate the mean attention entropy (c.f. Equation~\eqref{eq:entropy}) and analyze it by layer/head (Figure~\ref{fig:entropy-all-domains}), as well as by layer alone (Figure~\ref{fig:entropy-by-layer-excl-first-tok}) and head alone (Figure~\ref{fig:entropy-by-head-excl-first-tok}) across all four domains considered in this study: general web data, human-human conversations, code, and math.

In Figure~\ref{fig:entropy-all-domains},  the heatmaps represent the entropy by layer/head for web, human-human conversations, code, and math domains. Attention dispersion is highest in the human-human conversations domain. This is consistent with the attention distance difference plot in Figure~\ref{fig:attention-distance-difference-comparison}.  From layer $22$ to layer $36$, entropy is typically lower for web, code, and math domains; however, the entropy is high in multiple heads in these layers for human-human conversations. In the conversation domain, for each token, the model has to attend strongly to more tokens than in the rest of the domains -- this indicates higher complexity for the domain, which leads to higher attention dispersion in the model while understanding that domain. It also suggests that the model is less familiar with the human-human conversation domain, which can be explained by the scarcity of training data in the domain distribution (c.f. Section~\ref{conversation-data-in-web}). For web, code, and math domains in comparison, the entropy is noticeably lower. This indicates that the model has a more robust understanding of these domains, and the model can find an optimal attention strategy, reducing attention dispersion, which can be explained by the considerable amount of data from these domains that is reflected in the model's pretraining data (c.f. Section~\ref{language-model-llama2}). 

We also plot the entropy by removing attention to the first token in sequence, by layers and heads separately, shown in Figure~\ref{fig:entropy-by-layer-excl-first-tok} and Figure~\ref{fig:entropy-by-head-excl-first-tok} respectively. We remove the first token's entropy because we find that the model adds redundant attention to the first token which leads to high entropy, especially in the first layer.\footnote{The plots without removing attention to the first token are shown in Figure~\ref{fig:entropy-by-layer} and Figure~\ref{fig:entropy-by-head} in the Appendix.}  As shown in Figure~\ref{fig:entropy-by-layer-excl-first-tok}, certain layers -- specifically layers $27$ and $29$, which are mid-layers of the model -- have significantly higher mean attention entropy compared to the rest of the layers for the human-human conversation domain. A similar pattern can be seen in Figure~\ref{fig:entropy-by-head-excl-first-tok}, where heads $13$, $25$, and $38$ have high entropy. For web, code, and math domains, such high entropy is not exhibited by the model, indicating that the model has less familiarity with complex relationships in the human-human conversation domain as compared to others.

\begin{table}[!ht]
  \centering
  \begin{tabular}{lll}
    \toprule
    \cmidrule(r){1-2}
    Domain     & Overall Attention Entropy   \\
    \midrule
    Web &  $0.0083$ \\
    Human-Human Conversations & $0.0098$    \\
    Code & $0.0083$   \\
    Math & $0.0085$    \\
    \bottomrule
    \smallskip
  \end{tabular}
  \caption{Overall attention entropy averaged across heads and layers for domains.}
  \label{table:overall-attention-entropy-by-domain}
\end{table}

\subsubsection{Attention Interdependency Analysis}

We perform interdependency analysis between human-human conversations and other domains to gain deeper insight into the underlying structures that a model needs to utilize across these domains. For human-human conversations and web data, we perform analysis using theme segmentation and dependencies between themes within a conversation or document, as well as a token-level analysis. We only perform token-level interdependency analysis for code and math data, as thematic analysis on these domains does not provide much insight due to their logical and rule-driven nature. We calculate the average attention matrix across all samples in the domain averaged across all attention heads from the middle layer. 

We calculate $IF$ for these domains to understand the overall interdependency between tokens with 512 tokens in each sequence to get a quantitative evaluation of the interdependency by domain in Table \ref{table:IF-by-domain}.

\begin{table}[!ht]
  \centering
  \begin{tabular}{lll}
    \toprule
    \cmidrule(r){1-2}
    Domain     & IF   \\
    \midrule
    Web & $100.207$   \\
    Human-Human Conversations     & $141.869$    \\
    Code     & $106.466$    \\
    Math     & $110.848$    \\
    \bottomrule
    \smallskip
  \end{tabular}
  \caption{Interdependency Factor by domain, calculated for $N = 512$ across all samples.}
  \label{table:IF-by-domain}
\end{table}

\begin{figure}[!ht]
    \centering
    \includegraphics[width=1\linewidth]{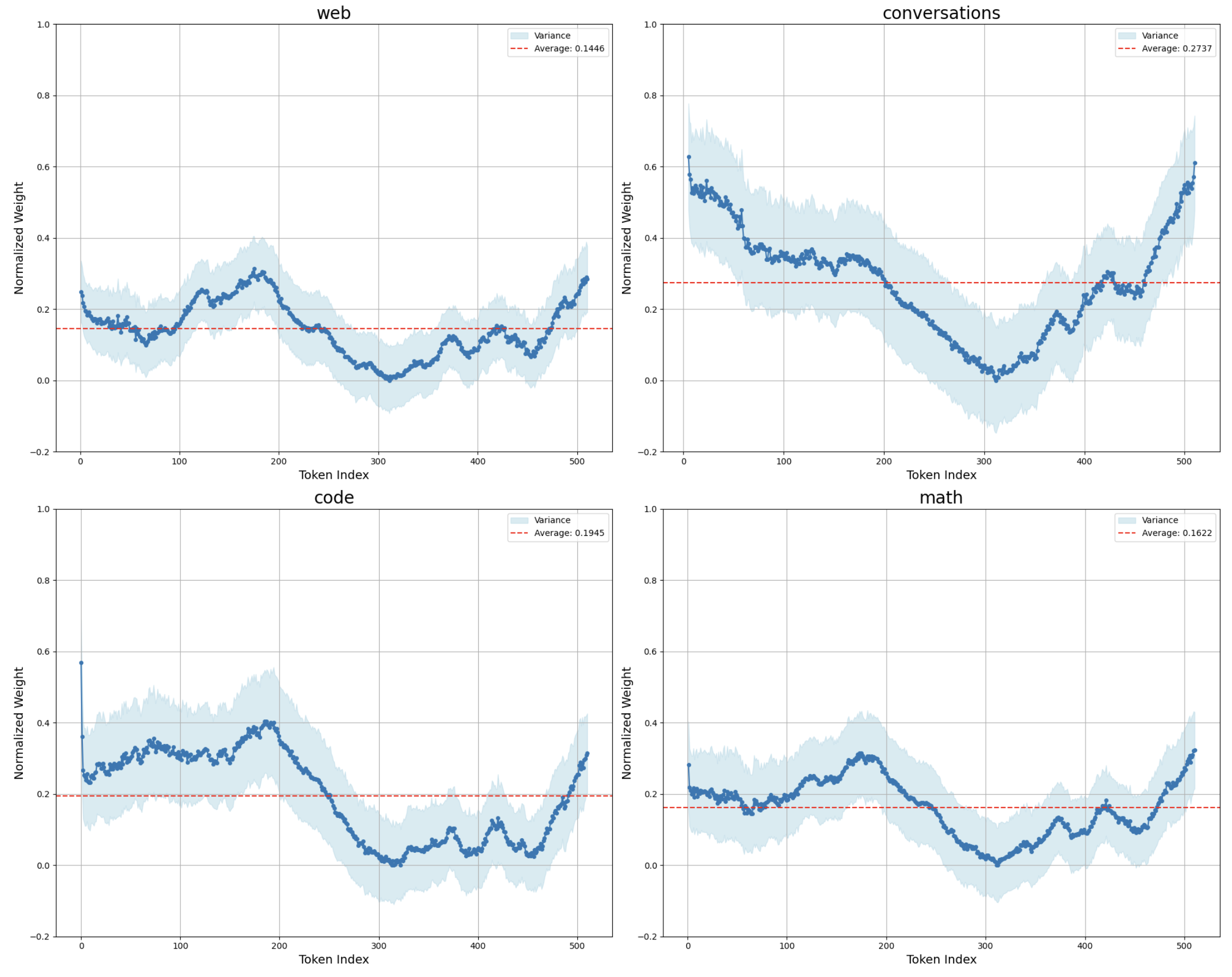}
 \caption{Normalized attention weights of interdependencies aggregated by the token for web, human-human conversations, code, and math domains respectively.}
    \label{fig:weights-by-domain}
\end{figure}

To further analyze the overall attention at each token, the individual token's weights can be calculated by aggregating the attention weights of the token towards the rest of the tokens as shown in Figure \ref{fig:weights-by-domain}. This provides us insights into how attention patterns change across the text sequence by domain. Fluctuations in weights by token index signify frequent changes in overall attention strength, and the value of weight indicates the overall strength of attention which is aggregation of attention values of token attending to all other tokens. Note that due to aggregation, the plot in Figure \ref{fig:weights-by-domain} no longer provides us with information about the global or local interdependencies. Human-human conversations require higher and longer attention resulting in higher average weight as compared to the rest of the domains. Due to more localized dependencies and rare long dependencies, we see that the average weight for web data is the lowest. The code data pattern is quite similar to the web except in the initial part of the sequence where higher attention can be seen as compared to the web with a higher average weight than the web. Similar trend can be observed in math domain.

\subsection{Language Model Representation}
\begin{figure}[!ht]
    \centering
    \includegraphics[width=1\linewidth]{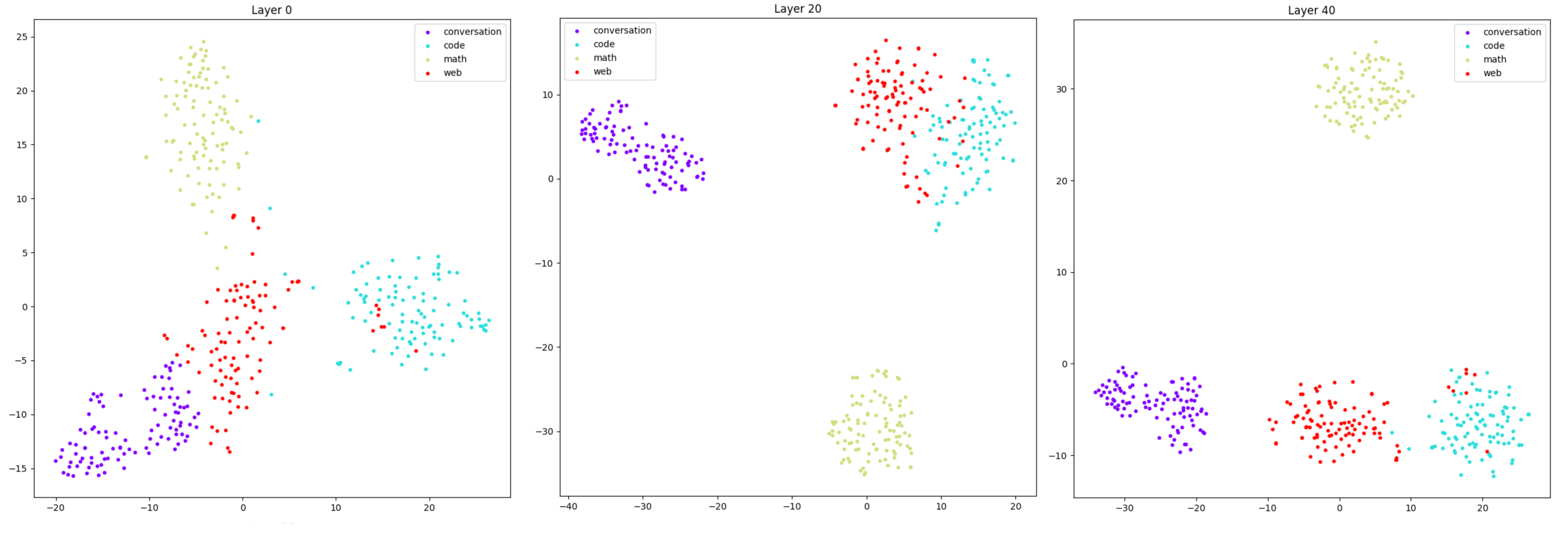}
    \caption{t-SNE plot of first, middle, and last layer of LLaMa-2 13b model by domains. Semantic characteristics of domains are different but closer in representation, however in deeper - middle, and last layers, human-human conversations, and math data are clearly represented significantly differently. While code representation is separate in the last layer, the middle layer shows there is quite an overlap in general data on the web and code in the middle layer representation.}
    \label{fig:tsne-representation}
\end{figure}

To understand the representation of language models by domains, we use t-SNE \cite{van2008visualizing} to visualize and compare the hidden state representations of the first, middle, and last layers of a general language model. We use the pretrain version of the LLaMa-2 13b model \cite{touvron2023llama} to analyze the learning distribution of a language model to have generality maintained. Early layers in language models learn the syntactic and semantic relationships in the sequences, and with deeper layers complex relationships are modeled capturing abstract and higher level understanding \cite{jawahar2019does, hao2021self}. In Figure \ref{fig:tsne-representation}, the representation of the first layer for domains is relatively closer to each other than the rest of the layers, with clear boundaries in the clusters, given that the semantic features of most of the language-based domains are similar but slightly nuanced. The middle layer representation shows some overlap in web and code data, but a clear and quite significant distance between human-human conversations and math data. We continue to see a similar pattern in the last layer with slightly better separation in web and code while conversations and math data continue to be distant from web and code domains. A language model trained on a general corpus along with data from various domains shows that the domain-specific learnings converge differently in a language model.

\section{Qualitative Analysis}
\begin{figure}
    \centering
    \includegraphics[width=\linewidth]{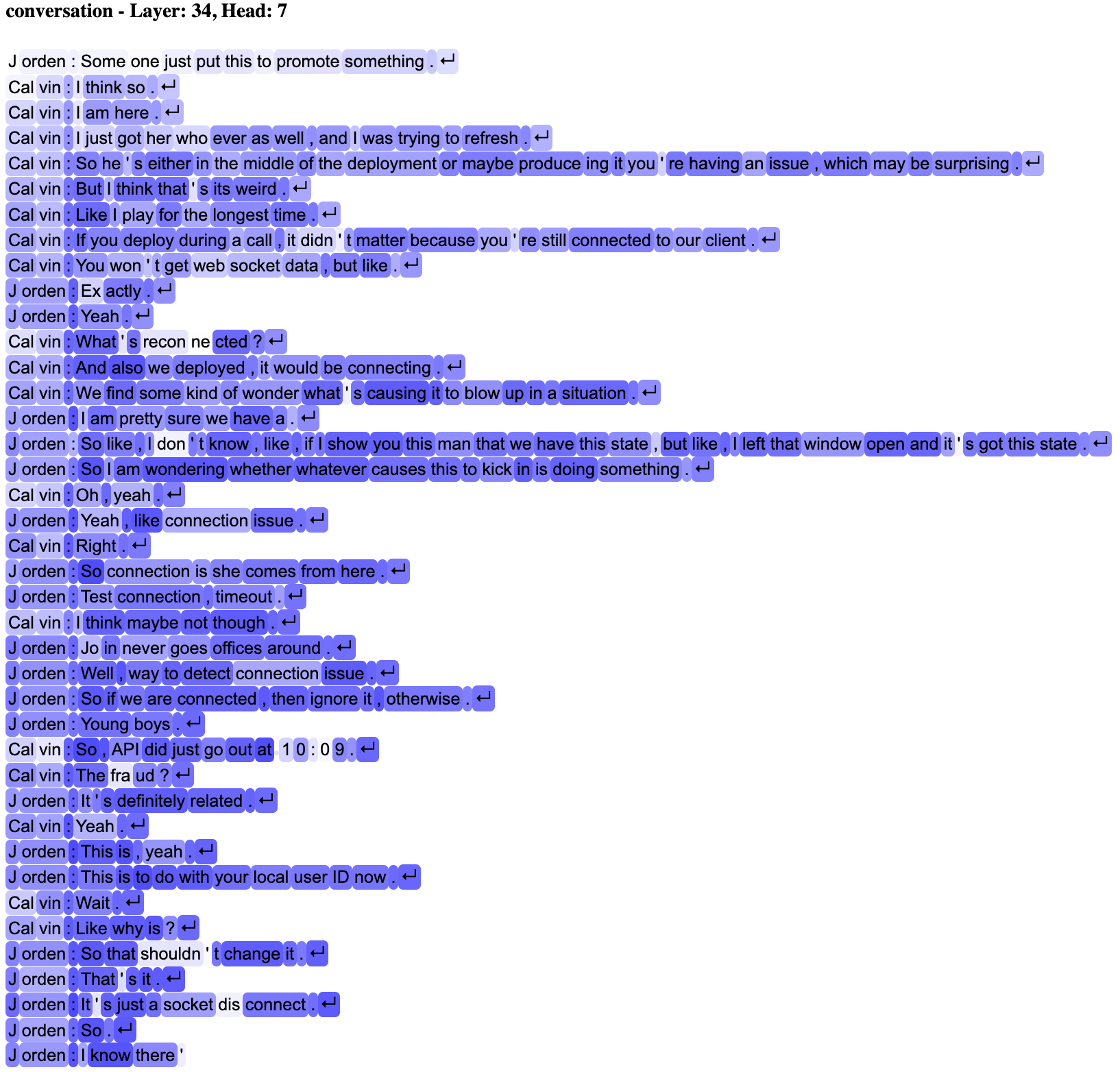}
    \caption{Human-Human Conversation example highlighted for mean attention entropy at each token for layer 34, head 7, showing the high attention diffusion compared to web, code, and math examples.}
    \label{fig:example-conv-entropy}
\end{figure}

\begin{figure}
    \centering
    \includegraphics[width=\linewidth]{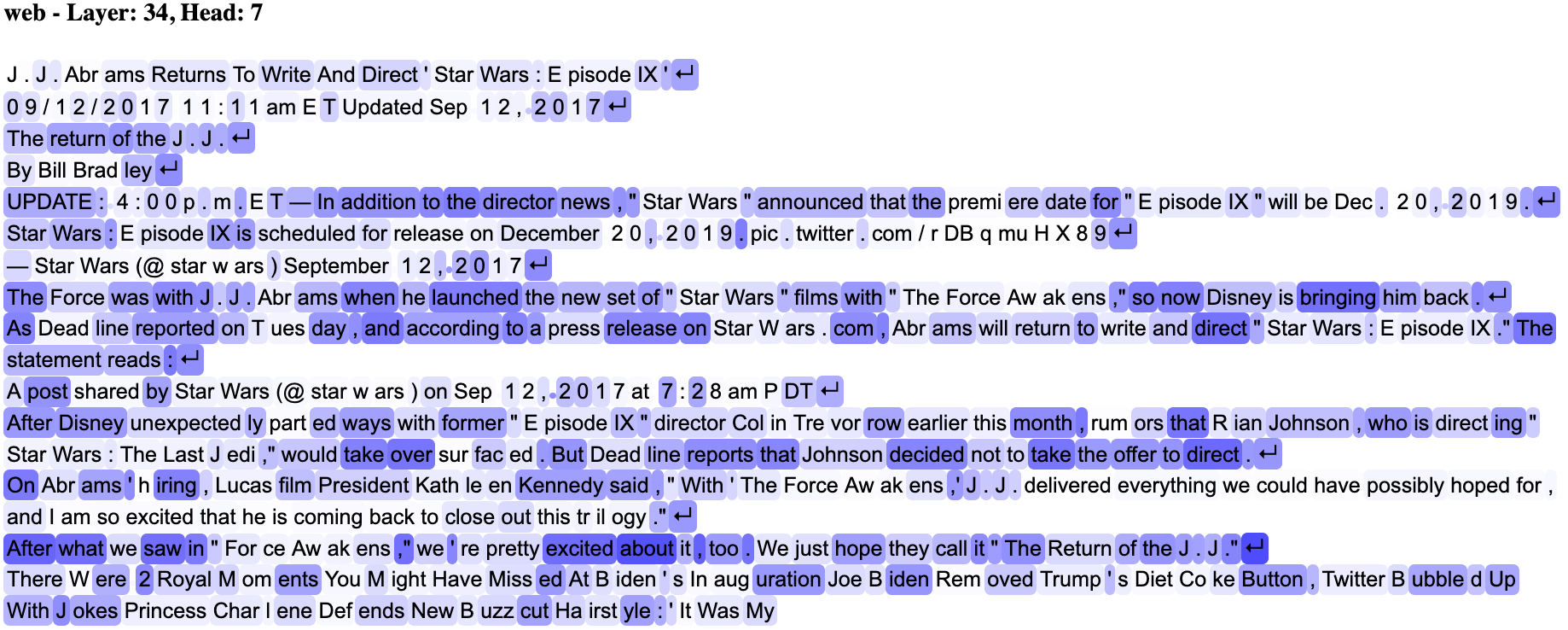}
    \caption{Web data example highlighted for mean attention entropy at each token for layer 34, head 7. Attention diffusion is significantly lower as compared to the human-human conversation as shown in Figure~\ref{fig:example-conv-entropy}.}
    \label{fig:example-web-entropy}
\end{figure}

\begin{figure}
    \centering
    \includegraphics[width=0.5\linewidth]{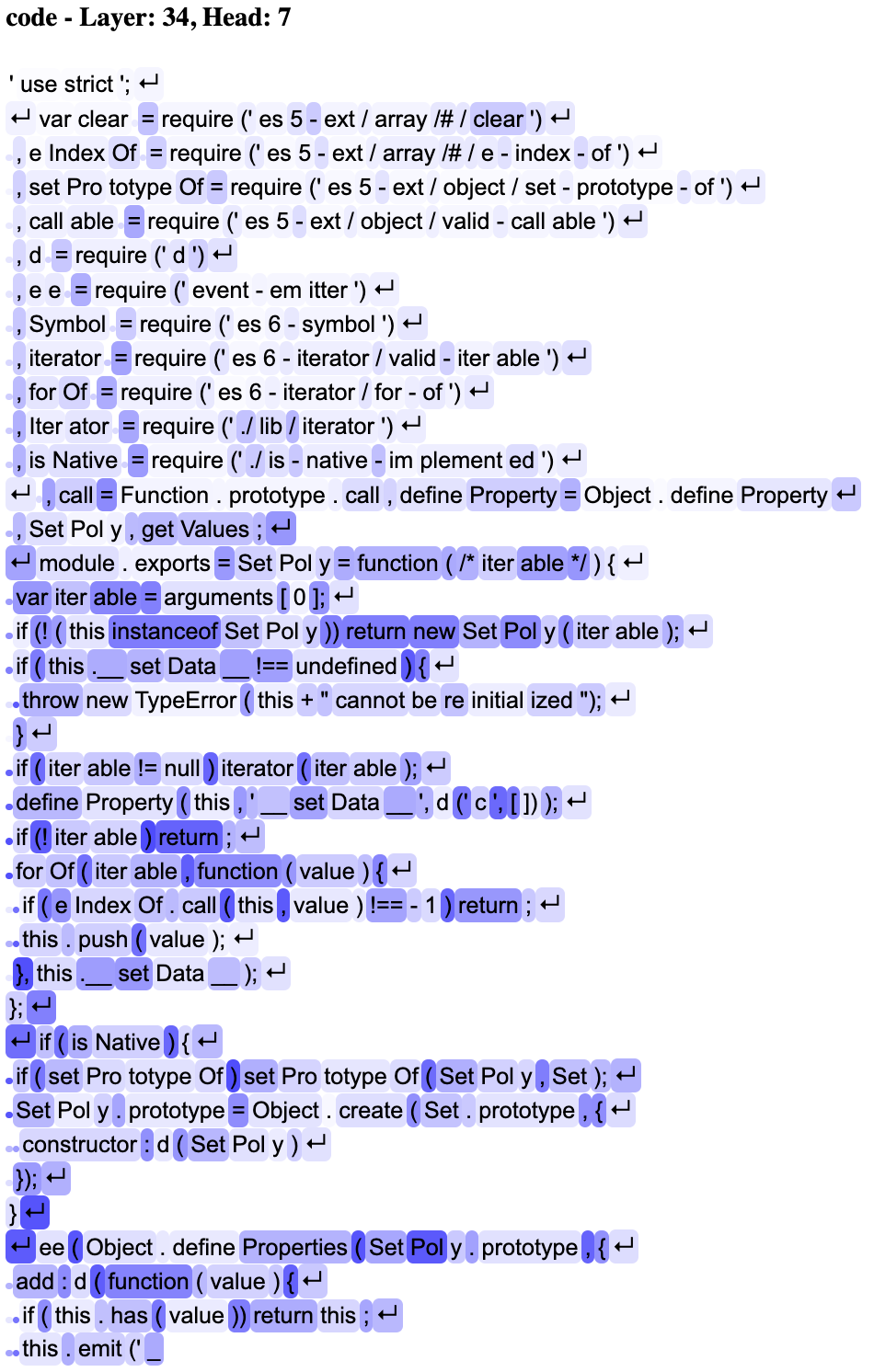}
    \caption{Code example highlighted for mean attention entropy at each token for layer 34, head 7. Attention diffusion is significantly lower as compared to the human-human conversation as shown in Figure~\ref{fig:example-conv-entropy} and equivalent to the web example in Figure~\ref{fig:example-web-entropy}.}
    \label{fig:example-code-entropy}
\end{figure}

\begin{figure}
    \centering
    \includegraphics[width=\linewidth]{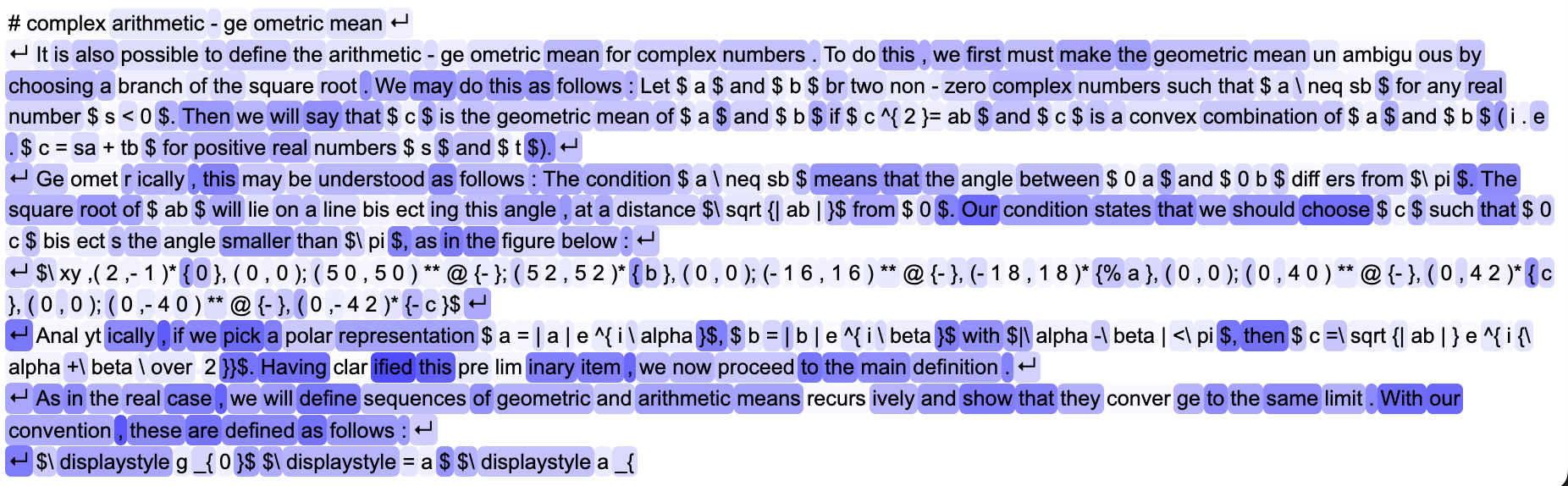}
    \caption{Math example highlighted for mean attention entropy at each token for layer 34, head 7. Attention diffusion is significantly lower as compared to the human-human conversation as shown in Figure~\ref{fig:example-conv-entropy} and equivalent to the web example in Figure~\ref{fig:example-web-entropy}.}
    \label{fig:example-math-entropy}
\end{figure}

To get an intuitive sense of patterns exhibited by specific layers and attention heads, we used a few examples from each domain to visually inspect the attention dispersion at each token in the example. Examples for each domain showing the attention entropy at each token in the example are shown in Figures \ref{fig:example-conv-entropy}, \ref{fig:example-web-entropy}, \ref{fig:example-code-entropy}, and \ref{fig:example-math-entropy}.  Several heads across layers show similar attention dispersion across all domains, however, certain heads show higher attention dispersion in human-human conversations as compared to other domains. We find that the initial layers have high entropy across all domains, whereas the middle and last layers have relatively high entropy in human-human conversations, as compared to web, code, and math domains also indicated by the attention dispersion analysis results.
\section{Discussion}

Our analysis reveals insights into the nuanced dynamics of natural human conversations as contrasted with structured forms of data like web content, code, and maths. The analysis supports the hypothesis that human conversations exhibit a unique set of characteristics, which necessitates a more robust modeling of long-term contextual relationships. This is indicative of the inherent nature of human dialogue, where meaning often unfolds cumulatively across exchanges, requiring an extended contextual memory from models.

The findings from attention distance and dispersion analysis reveal the unique demands that human conversation imposes on language models. The pronounced differences in deeper layers when comparing conversations to web data suggest that modeling of abstract long-term contextual relationships is more prevalent in human conversations. In contrast, the structural dependencies inherent in code and the balanced attention demands in mathematical texts present a different set of challenges for attention mechanisms, emphasizing the versatility required of language models to adapt across domains.

Furthermore, the study highlights the complexity of attention mechanisms in processing conversations. The higher attention entropy observed in human conversations suggests a need for models to adopt a broader focus, capturing the necessary context and nuances. This contrasts sharply with more structured domains, where attention can be more narrowly directed towards key tokens and patterns. This indicates a higher complexity of the domain and less definitive understanding of the domain on the part of the model.

The interdependency among tokens -- which is particularly pronounced in human conversation data -- hints at the rich, interconnected structure of human communication. This reflects the multifaceted nature of human communication where themes, emotions, long-term contexts, and unspoken cues intertwine deeply.

Additionally, our exploration through t-SNE visualizations illustrates how domain-specific learnings are distinctly encoded within models. The visual distinction across the domains being compared when it comes to the model's deeper layers underscores the capacity of models to differentiate and specialize based on the unique demands of each domain: a crucial aspect for improving language model performance across diverse linguistic tasks.

These findings underscore the importance of incorporating diverse and nuanced data into the training of language models, and of training models specialized on human conversations to better capture their inherent complexity. The distinct characteristics of conversations, as highlighted through our analyses, suggest that models require further refinement to accurately understand and generate human-like dialogue. This includes enhancing models' ability to manage long-term dependencies, and to distribute attention in a manner that mirrors the dynamic nature of a conversation.

Finally, the observed differences across domains advocate for the development of more adaptable and flexible attention mechanisms within language models. Such mechanisms would enable models to shift their focus and processing strategies depending on the type of data being handled, potentially improving performance across a wider range of linguistic and cognitive tasks. Our analysis opens avenues for further research into the underlying cognitive and linguistic processes that these models emulate. By delving deeper into the interplay between attention mechanisms, domain specificity, and language understanding, we can uncover more about the nature of human cognition and communication.

\section{Conclusion}
In this study, we examined how transformer-based language models (using LLaMa-2 13b as a representative) process natural human conversations in contrast to web content, code, and mathematical texts. We found a general lack of sufficient representation of human-human conversations in web data, which is the largest constituent of pretraining data in most current large language models. Our findings highlight that human-human conversational data challenges a(ny) model into managing long-term contextual relationships, as well as to have more dependencies, especially in deeper layers. This effect is not as pronounced in structured data domains like code or math, and in better represented corpora like web data. Attention entropy tends to be higher in human-human conversations, indicating the need for a broader focus on the part of the model to grasp the nuanced and dynamic nature of human dialogue.

We find that while language models exhibit domain-specific attention patterns, there is a notable gap in their ability to process human conversation data effectively. For example, code and math data require a more focused attention mechanism due to their structured nature, which contrasts with the wide-ranging attention needed for understanding complex human conversations. Additionally, our interdependency analysis highlights the complex web of relationships present in conversational data, underscoring the intricacy of human communication.

Our analysis emphasizes the importance of domain specialization in language models to enhance their understanding and handling of human conversations; and indicates that training language models with a vast amount of high-quality authentic human conversations during the pretraining phase is an essential requirement in bridging the gap in model performance.

In conclusion, our analysis highlights the scarcity of diverse and authentic human conversation data, the higher complexity of human conversations as perceived by current language models, and the lack of specialization in today's language models toward natural human conversations. It highlights the need for domain-specialized language models for human conversations due to significant differences and gaps in attention behaviors exhibited by current general and generic language models, and indicates clear patterns in domain-specific modeling that can be exploited in the creation of language models that are specific to human conversations.

\bibliographystyle{unsrtnat}
\bibliography{bibliography/bibilography}

\clearpage

\section*{Appendix}

\begin{figure}[!ht]
    \centering
    \includegraphics[width=1\linewidth]{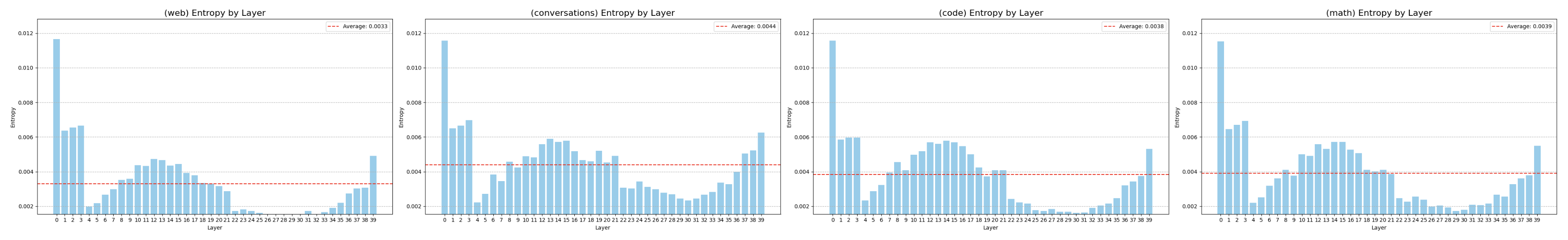}
    \caption{Mean attention entropy by layer without removing first token attention for web, human-human conversations, code, and math domains respectively.}
    \label{fig:entropy-by-layer}
\end{figure}

\begin{figure}[!ht]
    \centering
    \includegraphics[width=1\linewidth]{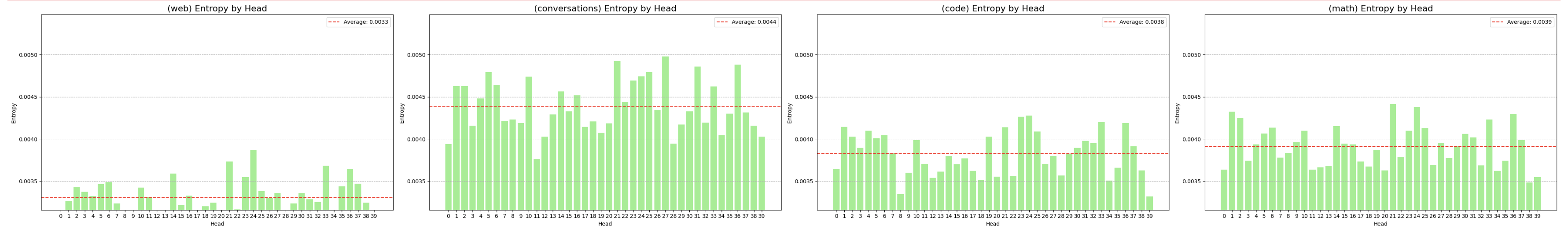}
    \caption{Mean attention entropy by head without removing first token attention for web, human-human conversations, code, and math domains respectively.}
    \label{fig:entropy-by-head}
\end{figure}

\end{document}